\documentclass{Interspeech}

\interspeechcameraready

\title{EmotionRankCLAP: Bridging Natural Language Speaking Styles and Ordinal Speech Emotion via Rank-N-Contrast}

\author[affiliation={1,2}]{Shreeram Suresh}{Chandra}
\author[affiliation={3}]{Lucas}{Goncalves}
\author[affiliation={4}]{Junchen}{Lu}
\author[affiliation={5}]{Carlos}{Busso}
\author[affiliation={1}]{Berrak}{Sisman}

\email{busso@cmu.edu, sisman@jhu.edu}

\affiliation{}{$^1$Center for Language and Speech Processing (CLSP), Johns Hopkins University}{USA}
\affiliation{}{\quad $^2$The University of Texas at Dallas, USA \quad $^3$Amazon, USA  \quad$^4$NUS}{Singapore}
\affiliation{}{ $^5$Language Technologies Institute (LTI), Carnegie Mellon University}{USA}

\usepackage{booktabs}
\usepackage{graphicx} 
\usepackage{lipsum}   
\usepackage{caption}
\usepackage{fontawesome} 
\usepackage[most]{tcolorbox}
\usepackage{lipsum}    
\usepackage{amsmath}
\usepackage{cite}
\usepackage{hyperref}
\tcbset{
    professionalbox/.style={
        colframe=black,         
        colback=gray!10,        
        coltext=black,          
        sharp corners,          
        boxrule=0.5pt,          
        left=5pt,               
        right=5pt,              
        top=5pt,                
        bottom=5pt,             
        before skip=10pt,       
        after skip=5pt,         
        enhanced,               
    }
}

\begin{document}
\maketitle

\begin{abstract}
    Current emotion-based \emph{contrastive language-audio pretraining} (CLAP) methods typically learn by naïvely aligning audio samples with corresponding text prompts. Consequently, this approach fails to capture the ordinal nature of emotions, hindering inter-emotion understanding and often resulting in a wide modality gap between the audio and text embeddings due to insufficient alignment. To handle these drawbacks, we introduce EmotionRankCLAP, a supervised contrastive learning approach that uses dimensional attributes of emotional speech and natural language prompts to jointly capture fine-grained emotion variations and improve cross-modal alignment. Our approach utilizes a Rank-N-Contrast objective to learn ordered relationships by contrasting samples based on their rankings in the valence-arousal space. EmotionRankCLAP outperforms existing emotion-CLAP methods in modeling emotion ordinality across modalities, measured via a cross-modal retrieval task.\footnote{\href{https://kodhandarama.github.io/emotionrankclap.github.io/}{https://kodhandarama.github.io/emotionrankclap.github.io/}}

\end{abstract}

\section{Introduction}
    





The expression and perception of human emotion are inherently continuous in nature~\cite{russell2003core}. Emotions also possess an ordinal nature, as humans are more adept at detecting relative changes in expression rather than identifying absolute emotional states~\cite{yannakakis2018ordinal}. However, existing paralinguistic models that attempt to capture the ordinality of speech emotion primarily rely on dimensional attribute annotations~\cite{parthasarathy2017ranking,martinez2014don}, which limit their ability to fully represent the nuanced structure of emotional expression. We believe that fine-grained and ordinal nature of speech emotion can be more effectively captured with natural language descriptions.

Natural language supervision has emerged as a promising approach for enhancing audio and speech understanding. In particular, \emph{contrastive language-audio pretraining} (CLAP)~\cite{elizalde2023clap} has gained popularity as a method for aligning audio with natural language prompts. By sharing a common representation space across modalities, CLAP enables tasks such as zero-shot captioning \cite{10448115}, classification \cite{ghosh2024reclap}, and cross-modal retrieval \cite{deshmukh23_interspeech}. 

CLAP has also been adopted extensively for emotion tasks including \emph{speech emotion recognition} (SER)~\cite{pan2024gemo}, emotional \emph{text-to-speech} (TTS)~\cite{jing2024enhancing} and \emph{emotion audio retrieval} (EAR)~\cite{dhamyal2024prompting}. GemoCLAP~\cite{pan2024gemo} focuses on building a discriminative representation space for SER using categorical labels. ParaCLAP~\cite{jing24b_interspeech} and CLAP with prompt-augmentation~\cite{dhamyal2024prompting} improve supervision by describing acoustic properties of emotional audio. The work most similar to ours, CLAP4emo~\cite{lin2024clap4emo}, generates pseudo-captions using pre-trained \emph{large language models} (LLMs) based on categorical emotion annotations of speech utterances. With the current approach of using only categorical emotions, intra-class variability is overlooked—for instance, all speech-text pairs labeled as ``happiness'' are treated identically, ignoring differences in intensity or expression. Likewise, inter-class relationships are not captured, such as the fact that ``disgust'' and ``fear'' are more closely related than ``happiness'' and ``fear''.


A key limitation in existing CLAP-based models is their reliance on the diagonal-constraint-based \emph{symmetric cross-entropy} (SCE) loss ~\cite{radford2021learning}, which presents two major drawbacks. First, at the batch level, this loss function fails to capture inter-emotion relationships across modalities. Since emotions are inherently ordinal, aligning each speech-text pair in isolation overlooks the structured relationships between different emotional states. Secondly, while emotion-based CLAP models leverage emotion annotations in text prompt design, they retain the loss formulation of CLIP~\cite{radford2021learning}, designed originally for self-supervised training, leading to a modality gap between text and audio embeddings at the end of training. Here, the modality gap refers to the insufficient overlap of embedding spaces of different modalities, a well-documented issue in cross-modal learning frameworks~\cite{yaras2024explaining}. We argue that this modality gap can be effectively reduced in a supervised setting by incorporating dimensional emotional attributes in speech.

To address these limitations, we adopt Rank-N-Contrast~\cite{zha2024rank}, a contrastive learning objective specifically designed to learn ordered representations by ranking samples relative to their positions in the target label space. This objective ensures that the learned representations maintain the intended ordinal structure, aligning with the target rankings. While extensively studied in regression tasks, its application to cross-modal representation learning, particularly for capturing the ordinality of emotions, remains unexplored. In this work, we introduce EmotionRankCLAP, a novel supervised contrastive learning strategy that uses dimensional emotional attributes to learn a continuous emotion embedding space with the cross-modal formulation of Rank-N-Contrast. Our key contributions are as follows:

\begin{itemize}
    \item We propose leveraging the ordinal nature of emotions to learn a fine-grained emotion embedding space, using the Rank-N-Contrast objective;
    \item We show that using Rank-N-Contrast as an alternative to symmetric cross entropy loss improves cross-modal alignment, bringing the distributions of the audio embeddings and text embeddings closer together;
    
    \item We formulate a cross-modal retrieval task that checks the emotion ordinal consistency of the audio and text embeddings -  and we show EmotionRankCLAP outperforms other emotion-based CLAP models in this test.
    \item We generate and release natural-language emotional speaking style descriptions based on dimensional emotion attributes from the MSP-Podcast corpus \cite{lotfian2017building} (release 1.12) to bridge the speech and text modalities in the CLAP model.
    
    
\end{itemize}


To the best of our knowledge, this study is the first to leverage the ordinal nature of speech emotions to align the continuums of dimensional speech emotion and natural language speaking style descriptions.

\section{Related work}


\subsection{Cross-modal contrastive  learning}
Contrastive learning has proven to be an effective approach for aligning multiple modalities in shared representation spaces~\cite{radford2021learning, elizalde2023clap, wu2023large}.
While unsupervised contrastive learning relies solely on modality co-occurrence, it can lead to imprecise alignments without capturing task-specific semantic relationships, prompting the exploration of supervised settings~\cite{khosla2020supervised, stewart2024emotion}. By incorporating supervision, contrastive learning frameworks can better capture fine-grained inter-modality relationships, making them particularly effective for emotion-related tasks. 
 Inspired by these strategies, we propose a cross-modal version of Rank-N-Contrast, leveraging dimensional emotional attributes as an additional supervision signal to improve speech-text alignment.





 \subsection{Natural language description of speech emotion}
 Emotion annotations have traditionally been limited to manually annotated categorical labels or dimensional attributes. However, recent advancements have shifted the focus towards using natural language, allowing for more descriptive representations of speech emotion.
This has been made possible thanks to caption generation capability of LLMs~\cite{kim2019audiocaps}. This capability has been adapted into multimodal SER models~\cite{dutta2025llm, 10848758} to generate pseudo-captions. Speech language models like SECap~\cite{xu2024secap}  and AlignCap~\cite{liang2024aligncap} present a paradigm shift away from SER and towards speech emotion captioning via speech language models. Similarly, emotional TTS models are increasingly prioritizing controllability by using natural speaking style prompts rather than relying solely on categorical emotion labels~\cite{yang2024instructtts}. Our approach leverages an LLM to generate speaking style descriptions in the absence of speech datasets with captions.

\section{EmotionRankCLAP}
We propose EmotionRankCLAP, a supervised cross-modal contrastive learning framework to align emotional speech with natural language speaking style descriptions in a shared embedding space, leveraging the ordinal nature of speech emotions through a Rank-N-Contrast learning objective.
\subsection{Problem Formulation}

Let \(\{X_i^a, X_i^t\}\) for \(i \in \{1, ..., N\}\) be a batch of $<$speech, text$>$ pairs. Input from audio and text modalities are first encoded via two separate encoders, \(f^a(.)\) and \(f^t(.)\), yielding embeddings:
\vspace{-1mm}
\[
\hat{X}^a_i = f^a(X^a_i); \quad \hat{X}^t_i = f^t(X^t_i),  \tag{1}
\]

where \(\hat{X}^a \in \mathbb{R}^{N\times V}\) and \(\hat{X}^t \in \mathbb{R}^{N\times U}\). We employ a pre-trained, frozen WavLM-based dimensional SER model.\footnote{\href{https://huggingface.co/3loi/SER-Odyssey-Baseline-WavLM-Multi-Attributes}{\scriptsize{https://huggingface.co/3loi/SER-Odyssey-Baseline-WavLM-Multi-Attributes}}
}~\cite{goncalves2024odyssey}
as the audio encoder \( f^a(.) \), extracting 1024-dimensional embeddings via attentive statistics pooling across the temporal dimension from the last transformer layer. The text encoder \( f^t(.) \) is a pre-trained, frozen DistilRoBERTa model \footnote{\href{https://huggingface.co/j-hartmann/emotion-english-distilroberta-base}{\scriptsize{https://huggingface.co/j-hartmann/emotion-english-distilroberta-base}}}~\cite{hartmann2022emotionenglish}, using the final-layer [CLS] token as a 768-dimensional embedding. These representations are then projected to the same dimension \(D = 512\):
\vspace{-2mm}
\[
\hat{E}^a_i = proj^a(\hat{X}^a_i); \quad \hat{E}^t_i = proj^t(\hat{X}^t_i),  \tag{2}
\]
where \( \hat{E}^a, \hat{E}^t \in \mathbb{R}^{N \times D} \) are the projected embeddings, and \( proj^a \) and \( proj^t \) are modules with a linear transformation followed with ReLU activation. The goal of EmotionRankCLAP is to align the two modalities in the same embedding space while preserving ordinality to capture the dimensional nature of emotion in both text descriptions and speech.


\subsection{Supervised contrastive learning with Rank-N-Contrast}

Emotions are inherently continuous and ordinal, meaning that within any batch of emotional speech and its corresponding speaking style descriptions, a structured relationship exists between each possible pair, totaling
$N \times N$ cross-modal pairs. To learn this structured relationship,
we adopt Rank-N-Contrast, which contrasts samples based on their rankings in valence-arousal label space.

In the proposed formulation, we jointly model the ordinality of valence and arousal by considering them together in the label space. 
Valence reflects the sentiment expressed in the utterance, ranging from negative to positive. Arousal indicates the level of activation, with values spanning from calm to highly active.


For a given audio embedding anchor \(\hat{E}^a_{i}\), the likelihood of association with a text embedding \(\hat{E}^t_{j}\) depends on the relative distance of their labels in the valence-arousal space. Emotional distance is assessed by the \(L_2\) distance between ($valence^a_{i}$, $arousal^a_{i}$) and ($valence^t_{j}$, $arousal^t_{j}$), where closer samples are considered more alike. Here, \(i\) and \(j\) denote sample indices. 

 Let \(S_{i,j} := \{\hat{E}^t_{k} \mid d(\hat{E}^a_{i}, \hat{E}^t_{k}) > d(\hat{E}^a_{i}, \hat{E}^t_{j}\))\} denote the set of text embeddings
that are of higher rank than \(\hat{E}^t_{j}\) in terms of label distance with respect to \(\hat{E}^a_{i}\), where \(d(\cdot, \cdot)\) is the \(L_{2}\) distance measure between two labels in the valence-arousal plane.

Then the normalized likelihood of \(\hat{E}^t_{j}\) given \(\hat{E}^a_{i}\) and \(S_{i,j}\) can be written as
\vspace{-3mm}
\[
P(\hat{E}^t_{j} \mid \hat{E}^a_{i}, S_{i,j}) = \frac{\exp(\text{sim}(\hat{E}^a_{i}, \hat{E}^t_{j}) / \tau)}{\sum_{\hat{E}^t_{k} \in S_{i,j}} \exp(\text{sim}\hat{E}^a_{i}, \hat{E}^t_{k}) / \tau)}, \tag{3}
\]
where \(S_{i,j}\) represents the set of all \(\hat{E}^t_k\) that satisfy the ranking condition with respect to \(\hat{E}^{a}_{i}\) and \(\hat{E}^{t}_{j}\). This set contains the corresponding negative pairs for the positive pair $\hat{E}^a_{i}, \hat{E}^t_{j}$.
The similarity function \( \text{sim}(x,y) = \frac{x^T y}{\lVert x \rVert \cdot \lVert y \rVert} \) calculates the cosine similarity between cross-modal features and \(\tau\) denotes the temperature parameter. Defining this objective over all samples in a batch, we get the Rank-N-Contrast cross- modal loss: \vspace{-2mm}
 \[
\mathcal{L}_{\text{RNC-CM}} 
= \frac{1}{N^{2}} \sum_{i=1}^{N} \sum_{j=1}^{N} 
- \log P(\hat{E}^t_{j} \mid \hat{E}^a_{i}, S_{i,j}).
\tag{4}
\]
The loss function \( \mathcal{L}_{\text{RNC-CM}} \) exploits the continuous structure of the valence-arousal label space to ensure that emotional speech samples and speaking style descriptions with similar valence-arousal values also remain close in the learned representation space. The Rank-N-Contrast formulation enhances cross-modal alignment by leveraging all 
$N \times N$ speech-text pairs within a batch to form positive-negative pairs based on a ranking criterion. 
Each positive pair is assigned corresponding negative pairs according to their similarity ranking, ensuring a structured contrastive learning process. In contrast, SCE uses only \(N\) positive pairs per batch, limiting cross-modal alignment.

\subsection{Illustrative example of positive/negative pair selection}
\begin{figure}[t!]
    \centering
    \includegraphics[width=0.42\textwidth]{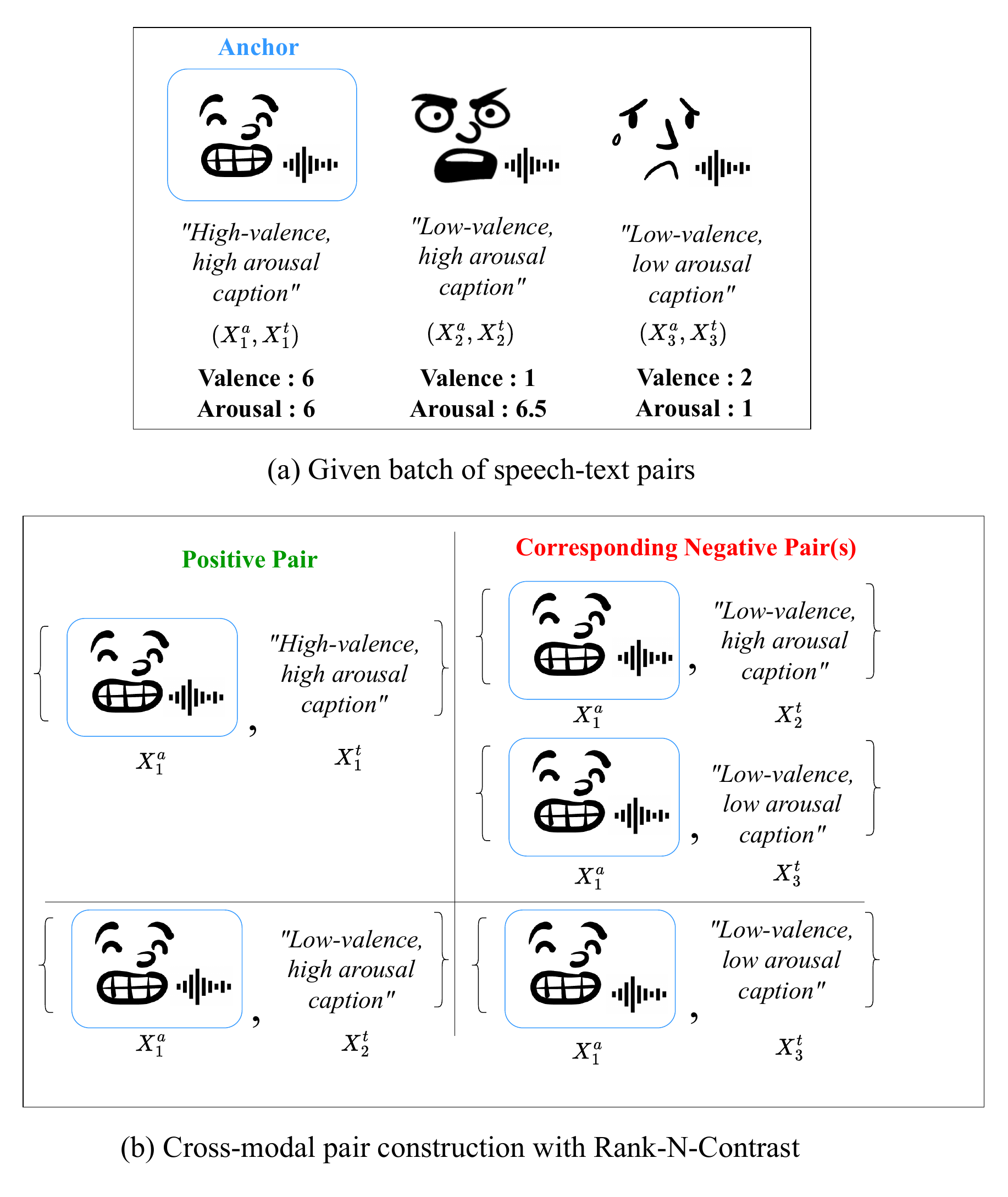} 
    \vspace{-4mm}
    \caption{Illustration of Rank-N-Contrast in a cross-modal setting. The anchor is boxed in blue. (a) A batch of speech-text pairs along with their valence-arousal labels.  (b) Positive and negative pair selection via Rank-N-Contrast criteria.}
    \vspace{-3mm}
    \label{fig:RNC} 
\end{figure}
We consider a batch of three speech-text pairs \( (X^a_i, X^t_i)\) (\( i \in \{1,2,3\} \)) with corresponding valence-arousal annotations as shown in  Figure~\ref{fig:RNC}(a).
As a demonstration of the positive/negative  pair selection, we set the first speech utterance \( X^a_1 \) as the anchor. Figure~\ref{fig:RNC}(b) illustrates two positive pairs and their corresponding negative pairs.

When considering the pair (\( X^a_1 \), \( X^t_1 \)) as positive, \( d(X^a_1, X^t_1) = 0 \) as both share the same label. This makes \( X^t_2 \) and \( X^t_3 \) negative samples since \( d(X^a_1, X^t_2) > 0 \) and \( d(X^a_1, X^t_3) > 0 \). Similarly, when \( X^t_2 \) forms a positive pair with \( X^a_1 \), \( X^t_3 \) is a negative sample since \( d(X^a_1, X^t_3) > d(X^a_1, X^t_2) \). In this case, \( X^t_1 \) is not a negative sample since \( d(X^a_1, X^t_1) < d(X^a_1, X^t_2) \).

Thus, structured relationships emerge: closer positive pairs tend to have more negative samples, reinforcing their closeness, while distant positive pairs have fewer negative samples, reducing their attraction. For a batch of \( N \), we iterate over each \( X^a_i \) for \( i = 1, \dots, N \), forming \( N \) relationships per anchor, resulting in \( N \times N \) structured relationships.


\vspace{-2mm}
\subsection{Generation of speaking style descriptions}
Existing speech emotion datasets are primarily designed for categorical and dimensional emotion recognition~\cite{busso2008iemocap,lotfian2017building}, providing annotations in terms of categorical labels and dimensional attributes (valence, arousal, dominance). In contrast, speech emotion captioning remains an emerging field, with limited datasets featuring manually annotated speaking style descriptions. To bridge this gap, we make use of an LLM~\cite{openai2023gpt4} to generate pseudo-captions based on valence and arousal. While this work focuses on these two attributes, incorporating dominance and other factors is left for future exploration. Figure \ref{fig:prompt_box} contains the prompt used to generate the natural language speaking style descriptions using OpenAI's o1 model.

\vspace{-3mm}
\begin{figure}[h!]
    \centering
    \begin{tcolorbox}[professionalbox, width=0.48\textwidth]
        \fontsize{6pt}{7pt}\selectfont\sffamily
        "Given the following scale of emotions - valence (1-very negative; 7-very positive), arousal (1-very calm; 7-very active), write a sentence describing a speaking style that is \{\textit{VALENCE}\} on valence, \{\textit{AROUSAL}\} on arousal. Do not use any numbers in the sentence. The sentence should start with: The person is speaking ..."
    \end{tcolorbox}
    \caption{Prompt used to generate emotional style descriptions based on valence-arousal values.}
    \vspace{-7mm}
    \label{fig:prompt_box}
\end{figure}

\section{Experiments}

In this section, we discuss the experimentation settings, the baselines and the evaluations used to probe the properties of the cross-modal embeddings.

\vspace{-2mm}

\subsection{Experimental setup}
\textbf{Dataset: }
We use the MSP-Podcast v1.12 corpus~\cite{lotfian2017building} for training, validation, and testing. Collected from real-world podcasts, it features significant acoustic variability, diverse speakers, and a broad range of emotional expressions, making it particularly challenging. We filter out samples with categorical emotion labels `X' (no agreement) and `O' (other), resulting in 90,022 training, 25,258 development, and 34,963 test samples (using only test 1 set). The large test set provides a comprehensive coverage of speaking styles. Each speech utterance is annotated with (valence, arousal, dominance) based on annotations provided by at least five annotators. We utilize the average score across annotators. \\
\textbf{Baselines: }
   


\begin{itemize}
    \item \textbf{CLAP-template:} This model is trained with the CLAP framework (SCE loss) using the text prompt: \textit``speech has \{categorical label\} emotion'' as input to the text encoder. 
    
    \item \textbf{CLAP4emo~\cite{lin2024clap4emo}:} This model replaces the pre-defined prompts in CLAP-template with natural language style descriptions generated with the help of ChatGPT ~\cite{openai2023gpt4} and an NRC lexicon~\cite{Mohammad13}. The captions for this model are generated following the pipeline described in their paper.
    
    \item \textbf{CLAP-SCE (A-V):} An ablation model trained with CLAP framework under SCE loss, where we use captions generated with dimensional emotional attributes instead of categorical emotion. The difference between this method and the proposed method is the loss function (SCE vs RNC). Here, (A-V) indicates that we use dimensional emotional attributes to generate the captions for this method.

    \item \textbf{SupConCLAP (A-V):} Another ablation baseline where we replace the SCE loss in CLAP-SCE (A-V) with SupCon~\cite{khosla2020supervised}, using categorical emotion labels to define the similarity matrix between text and audio embeddings.
     \item \textbf{ParaCLAP~\cite{jing24b_interspeech}:} This model is trained to align emotional audio with acoustic properties like pitch, jitter, shimmer, articulation rate using natural language descriptions. We extract acoustic information using Parselmouth-Praat~\cite{parselmouth}. 
\end{itemize}

\textbf{Training details: }
All input waveforms are resampled to 16\,kHz and cropped or zero-padded to 10 seconds. Text inputs are truncated to a maximum of 512 tokens. To ensure fair comparisons, all models share the same text and audio encoder architecture, with jointly trained projection layers. We train using the \textit{Adam} optimizer (learning rate \(1 \times 10^{-4}\)), a learnable temperature (initialized at 1.0), and a batch size of 64 on an NVIDIA L4 GPU. All models are implemented in PyTorch and trained for 15 epochs, selecting the checkpoint with the lowest validation loss.

\subsection{Evaluations and results}

\subsubsection{Cross-modal alignment}
\begin{table}[t]
    \centering
    \small
    \renewcommand{\arraystretch}{1.2}
    \setlength{\tabcolsep}{4pt}
    \caption{Comparison of methods on cross-modal alignment.\\ * denotes statistically significant improvement over all baselines (two-tailed p-test, p$<$0.05).}
    \vspace{-3mm}
    \begin{tabular}{lccc}
        \hline
        Method & MMD $\downarrow$ & Wass. Dist. $\downarrow$ \\
        \hline
        CLAP-SCE (A-V) &0.096$\pm$.002 & 0.180$\pm$.003 \\
        SupConCLAP (A-V) &0.4436$\pm$.003& 0.4172$\pm$.003 \\
        EmotionRankCLAP (A-V) & \textbf{*0.087$\pm$.001}& \textbf{*0.065$\pm$.007} \\
        \hline
        
    \end{tabular}
    
    \label{tab:cross_modal_alignment}
\end{table}
This evaluation tests the overlap of audio and text embedding spaces. We conduct 30 trials, each randomly sampling 5000 speech-text pairs from the MSP Podcast test-1 set. The audio embeddings are extracted from speech utterances, and text embeddings are extracted from the natural language descriptions. We measure \emph{maximum mean discrepancy} (MMD) \cite{gretton2012kernel} with a \emph{radial basis function} (RBF) kernel and Wasserstein distance \cite{peyre2019computational} both of which quantify alignment between the embedding distributions, where lower scores indicate better alignment. In this test, we only consider baselines which are trained with the same caption data as the proposed method, and we report the mean and standard deviation across the 30 trials.  As shown in Table~\ref{tab:cross_modal_alignment}, EmotionRankCLAP significantly outperforms the baselines by achieving the lowest MMD and Wasserstein distance scores, highlighting Rank-N-Contrast's superior cross-modal alignment compared to SCE and SupCon.



\subsubsection{Cross-Modality Emotion Ordinality Test}
\begin{table}[t]

    \centering
    \small
    \renewcommand{\arraystretch}{1.2}
    \setlength{\tabcolsep}{6pt}
    \caption{Comparison of cross-modal retrieval methods. * indicates a statistically significant improvement over all baselines (two-tailed p-test, p$<$0.05). AOC and VOC denote arousal and valence ordinal consistency, while KT represents Kendall's Tau.}
    \vspace{-3mm}
    \begin{tabular}{lcc}
        \hline
        Method &  AOC (KT) $\uparrow$ &  VOC (KT) $\uparrow$ \\
        \hline
        CLAP-template & 0.171$\pm$.26& 0.466$\pm$.14 \\
        CLAP4emo [10] & 0.284$\pm$.20 & 0.533 $\pm$.14\\
        ParaCLAP [8] &0.283$\pm$.20 & 0.217$\pm$.20 \\
        CLAP-SCE (A-V) &0.492$\pm$.15 & 0.505$\pm$.13\\
        SupConCLAP (A-V) & 0.346$\pm$.19& 0.530$\pm$.15\\
        EmotionRankCLAP (A-V)& \textbf{*0.552$\pm$.12} &\textbf{*0.616$\pm$.13} \\
        \hline
    \end{tabular}
    
    \label{tab:cross_modal_retrieval}
\vspace{-3mm}
\end{table}
In this evaluation, we examine how well the audio and text embedding spaces preserve ordinal consistency for dimensional emotional attributes. Specifically, ordinal consistency here means that speaking style descriptions indicating higher (or lower) valence (or arousal) should align more closely with speech utterances that exhibit correspondingly higher (or lower) valence (or arousal) levels. We design a cross-modal retrieval task to probe this property. 
Using the prompt in Figure~\ref{fig:prompt_box}, we generate 100 lists of speaking style descriptions, each containing 14 descriptions. We evaluate two properties: \emph{valence ordinal consistency} (VOC) and \emph{arousal ordinal consistency} (AOC). For VOC, we fix the arousal value in each list and vary valence from 0.5 to 7 in steps of 0.5. The fixed arousal value is incremented by 0.5 across lists, spanning the range [0.5,7], and resets to 0.5 after reaching 7. Conversely, for AOC, we fix the valence value in each list and vary arousal similarly. 
\begin{figure}[t!]
    \includegraphics[width=1.2\columnwidth]{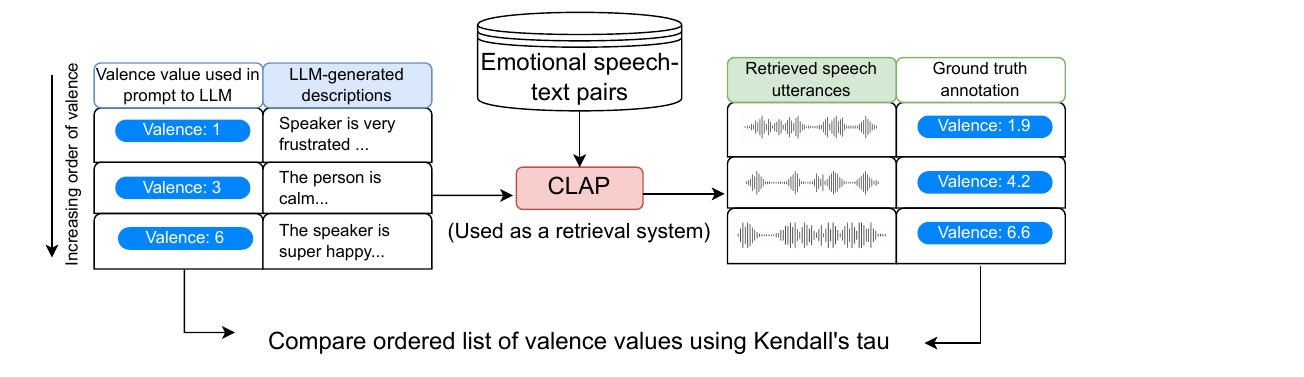} 
    \vspace{-7mm}
    \caption{Cross-Modality Emotion Ordinality Test: This figure shows a three-sample example for valence ordinal consistency, while the actual evaluation uses 14 samples per list, repeated across 100 lists for both valence and arousal.}
    \vspace{-2mm}
    \label{fig:ordinality}
    \vspace{-3mm}
\end{figure}
After generating these lists, we use the trained CLAP model as a retrieval system to find the most similar speech utterances for each textual description. The model encodes the speaking style prompt into a text embedding and retrieves the speech utterance with the closest audio embedding based on cosine similarity. We evaluate this property using the Kendall’s Tau coefficient (KT) \cite{kendall1938new}  between the valence (or arousal) values used to generate the speaking style descriptions and the valence (or arousal) values of the retrieved speech utterances, as shown in Figure~\ref{fig:ordinality}, and report the mean and standard deviation across 100 lists. To prevent redundant retrievals, each item is retrieved only once.

We observe that models trained using caption data generated with dimensional attribute guidance (denoted as (A-V)) are more consistent across both VOC and AOC tests.  This result highlights the importance of incorporating dimensional attributes when generating speaking style captions, as it helps in enhancing fine-grained cross-modal retrieval and in maintaining ordinal consistency. Interestingly, models trained with captions based on categorical emotions (CLAP-template and CLAP4emo) are competitive in VOC tests, but their performance degrades in AOC tests. Overall, EmotionRankCLAP achieves a significantly higher KT coefficient in both settings— VOC (lists with varying valence) and AOC (lists with varying arousal), as shown in Table. \ref{tab:cross_modal_retrieval}. This result demonstrates that our proposed cross-modal Rank-N-Contrast loss along with the use of captions generated with dimensional attribute guidance better preserves the ordinal structure of valence and arousal in the embedding spaces.
\section{Conclusions}
This work proposes EmotionRankCLAP, a supervised contrastive learning approach that leverages the ordinal nature of emotions to learn a cross-modal representation space to align dimensional speech emotions with corresponding speaking style descriptions. We generate natural language speaking style descriptions using dimensional attributes of speech emotion and we show that this is crucial in preserving emotion ordinality. We show that the proposed cross-modal formulation of Rank-N-Contrast loss improves cross-modal alignment between text and audio embedding spaces. We also design a cross-modal retrieval task to check ordinal consistency between the embedding spaces, and show that EmotionRankCLAP preserves ordinal nature of both valence and arousal better compared to other emotion-based CLAP models. In the future, we will explore other speech emotion tasks that take advantage of close cross-modal alignment and ordinal structure in the embedding space.
\vspace{-3mm}
\section{Acknowledgment}
This work is supported by NSF CAREER award IIS-2338979.

\bibliographystyle{IEEEtran}
\bibliography{mybib}

\end{document}